\documentclass[11pt,a4paper]{article}
\usepackage[hyperref]{acl2021}
\usepackage{times}
\usepackage{latexsym}
\usepackage{multirow}
\usepackage{graphicx}
\usepackage{booktabs}
\usepackage{amsmath}
\usepackage{xspace}

\usepackage{enumitem}
\setitemize[1]{itemsep=0pt,partopsep=0pt,parsep=\parskip,topsep=0pt,leftmargin=10pt}

\usepackage{microtype}

\aclfinalcopy 
\def\aclpaperid{1214} 

\usepackage{xr}
\makeatletter
\newcommand*{\addFileDependency}[1]{
  \typeout{(#1)}
  \@addtofilelist{#1}
  \IfFileExists{#1}{}{\typeout{No file #1.}}
}
\makeatother

\newcommand*{\myexternaldocument}[1]{
    \externaldocument{#1}
    \addFileDependency{#1.tex}
    \addFileDependency{#1.aux}
}
\myexternaldocument{appendix}

\newcommand\ourmodel{HyKnow\xspace}

\title{\ourmodel: End-to-End Task-Oriented Dialog Modeling with Hybrid Knowledge Management}

\author{
Silin Gao$^{1*}$, Ryuichi Takanobu$^{1*}$, Wei Peng$^{2}$, Qun Liu$^{2}$, Minlie Huang$^{1\dagger}$ \\
$^1$ CoAI Group, DCST, IAI, BNRIST, Tsinghua University, Beijing, China \\
$^2$ Huawei Technologies, Shenzhen, China \\
$^1$ {\tt gsl16@tsinghua.org.cn, gxly19@mails.tsinghua.edu.cn,} \\
{\tt aihuang@tsinghua.edu.cn} \\
$^2$ {\tt peng.wei1@huawei.com, qun.liu@huawei.com}
}

\date{}

\begin{document}
\maketitle
\begin{abstract}
Task-oriented dialog (TOD) systems typically manage structured knowledge (e.g. ontologies and databases) to guide the goal-oriented conversations.
However, they fall short of handling dialog turns grounded on unstructured knowledge (e.g. reviews and documents).
In this paper, we formulate a task of modeling TOD grounded on both structured and unstructured knowledge.
To address this task, we propose a TOD system with hybrid knowledge management, \ourmodel.
It extends the belief state to manage both structured and unstructured knowledge, and is the first end-to-end model that jointly optimizes dialog modeling grounded on these two kinds of knowledge.
We conduct experiments on the modified version of MultiWOZ 2.1 dataset, where dialogs are grounded on hybrid knowledge.
Experimental results show that \ourmodel has strong end-to-end performance compared to existing TOD systems.
It also outperforms the pipeline knowledge management schemes, with higher unstructured knowledge retrieval accuracy.
\end{abstract}
\renewcommand{\thefootnote}{\fnsymbol{footnote}}
\footnotetext[1]{Equal contribution.}
\footnotetext[2]{Corresponding author.}
\renewcommand{\thefootnote}{\arabic{footnote}}

\begin{figure}[t]
\centering
\includegraphics[width=1.0\columnwidth]{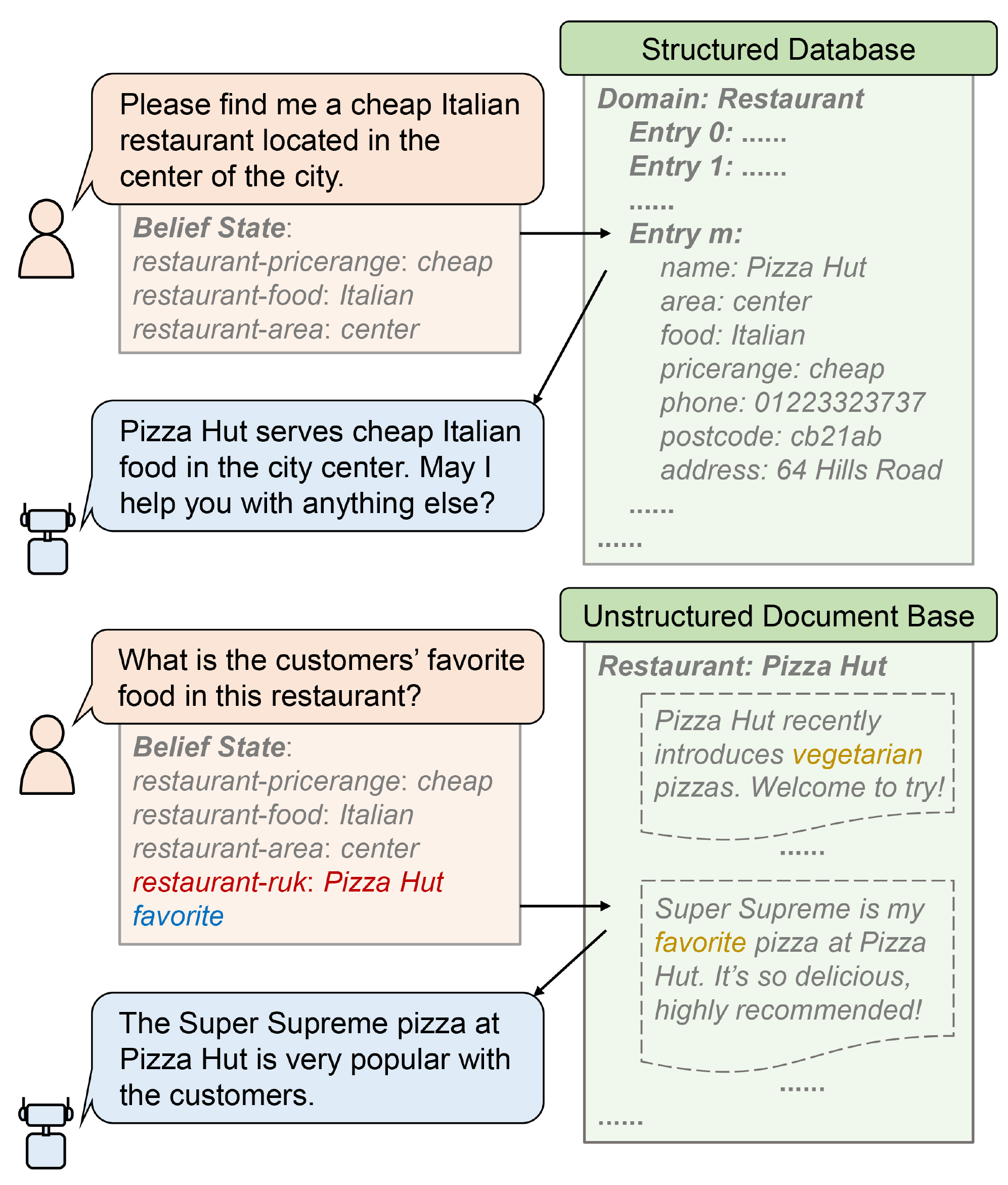}
\caption{Illustration of task-oriented dialog modeling with hybrid knowledge management. Words in red and blue illustrate the new domain-slot-value triple and the topic of user utterance that we introduce into the belief state, respectively. Words in yellow illustrate the topics of documents that we extract through preprocessing.}
\label{sskm}
\end{figure}

\section{Introduction}
Recently, Task-Oriented Dialog (TOD) systems \cite{mehri2019structured,zhang2020probabilistic,zhang2020task,le2020uniconv,hosseini2020simple,peng2020soloist,li2021multi} have achieved promising performance on accomplishing user goals.
Most systems typically query \textit{structured knowledge} such as tables and databases based on the user goals, and use the query results to guide the generation of system responses, as shown in the first dialog turn in Fig.~\ref{sskm}.

However, real-world task-oriented conversations often step into dialog turns which are grounded on \textit{unstructured knowledge} \cite{feng2020doc2dial}, such as passages and documents.
For example, as the second dialog turn in Fig.~\ref{sskm} shows, the user asks about customers' favorite food at \textit{Pizza Hut}, which is grounded on the customer reviews of this restaurant.
Current TOD systems fall short of handling such dialog turns since they cannot utilize relevant unstructured knowledge. This deficiency may interrupt the dialog process, causing difficulties in tracking user goals and generating system responses.

In this work, we consider incorporating more various forms of domain knowledge into the TOD systems.
Therefore, we define a task of modeling TOD whose turns involve either structured or unstructured knowledge.
In turns involving structured knowledge, the system needs to track the user goals as triples and use them to perform database queries, whose results are used to generate the system response.
While in turns involving unstructured knowledge, the system manages a document base to retrieve relevant references for generating the response.

To address our defined task, we propose a task-oriented dialog system with \textbf{Hy}brid \textbf{Know}ledge management (\ourmodel).
This model extends the belief state to handle TODs grounded on hybrid knowledge, and further uses the extended belief state to perform both database query and document retrieval, whose outputs are thereby used to generate the final response.
We consider two implementations of our system, with different schemes of extended belief state decoding. 
Both implementations are in an end-to-end multi-stage sequence-to-sequence (Seq2Seq) \cite{lei2018sequicity,liang2020moss,zhang2020probabilistic,zhang2020task} framework, where dialog modeling grounded on the two kinds of knowledge can be jointly optimized.

We evaluate our system on the modified version of MultiWOZ 2.1 \cite{kim2020beyond} dataset, where dialogs are grounded on hybrid knowledge. Experimental results show that \ourmodel outperforms existing TOD systems which do not leverage large pretrained language models, no matter whether they add extra unstructured knowledge management or not. It also has a higher accuracy in unstructured knowledge retrieval, compared to the pipeline knowledge management schemes.

Our contributions are summarized as below:
\begin{itemize}
    \item We formulate a task of modeling TOD grounded on both structured and unstructured knowledge, to incorporate more domain knowledge into the TOD systems.
    \item We propose a TOD system \ourmodel to address our proposed task. It extends the belief state to manage hybrid knowledge, and is the first end-to-end model to jointly optimize dialog modeling grounded on the two kinds of knowledge.
    \item Experimental results show that \ourmodel has strong performance in dialog modeling grounded on hybrid knowledge.\footnote{The code is available at \url{https://github.com/truthless11/HyKnow}}
\end{itemize}

\section{Related Work}
TOD systems usually use belief tracking, i.e. dialog state tracking (DST) to trace the user goals, i.e. \textit{belief states}, through multiple dialog turns \cite{williams2013dialog,henderson2014second}. The states are converted into a representation of constraints based on different schemes to query the databases \cite{el2017frames,budzianowski2018multiwoz,rastogi2020towards,zhu2020crosswoz}. The entry matching results are then used to generate the system response.

With the development of intelligent assistants, the system should have a good command of massive external knowledge to better accomplish complicated user goals and improve user satisfaction. To realize this, some researchers \cite{zhao2017generative,yu2017learning,akasaki2017chat} equip the system with chatting capability to address both task and non-task content in TODs. Other studies apply knowledge graph \cite{liao2019deep,yang2020graphdialog} or tables via SQL \cite{yu2019cosql} to enrich the knowledge of TOD systems. However, all these studies are still limited in dialog modeling grounded on structured knowledge.

There are a couple of studies to integrate unstructured knowledge into TOD modeling recently. \citet{kim2020beyond} introduce knowledge snippets to answer follow-up questions out of the coverage of databases. \citet{feng2020doc2dial} formulate document-grounded dialog for information seeking tasks. However, they only focus on dialog turns grounded on unstructured knowledge instead. In this paper, we aims to fill the gap of managing domain-specific knowledge with various sources and structures in traditional TOD systems.

\begin{figure*}[t]
\centering
\includegraphics[width=\textwidth]{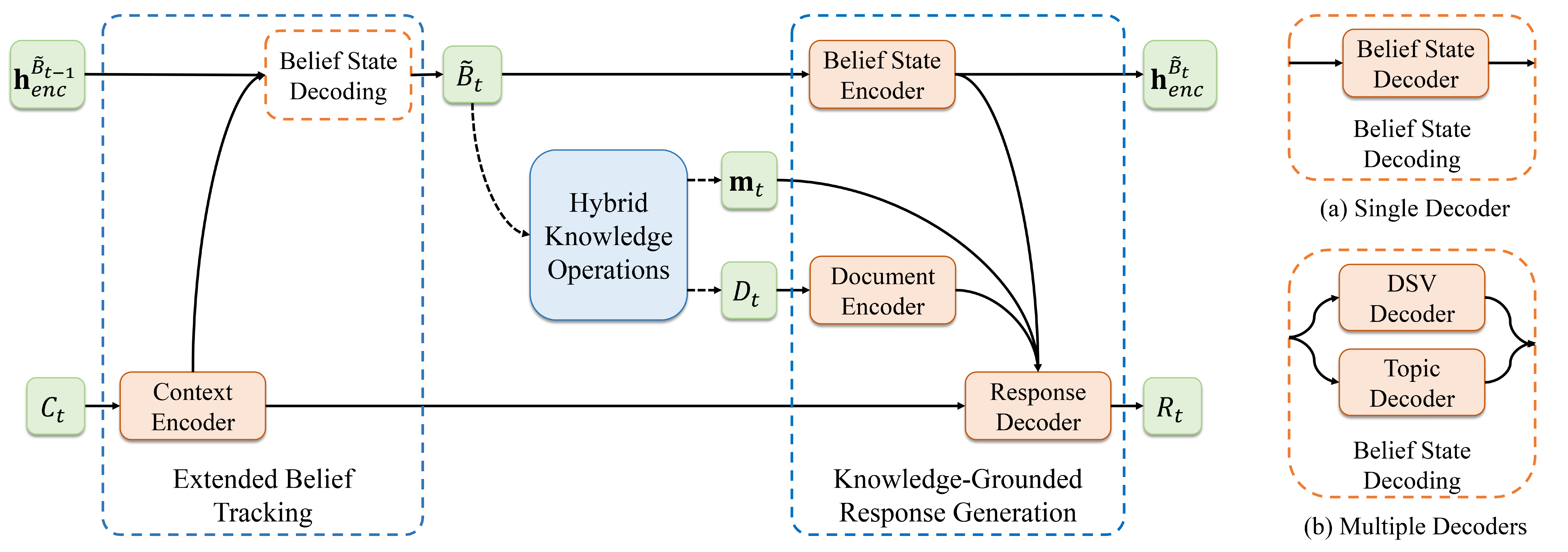}
\caption{Overview of \ourmodel. Solid arrows denote the input/output of the encoders or decoders. Dashed arrows denote the knowledge operations. $C_{t}$, $\textbf{m}_{t}$, $D_{t}$ and $R_{t}$ represent turn $t$'s dialog context, DB query result, relevant document and system response. $\widetilde{B}_{t}$ and $\textbf{h}_{enc}^{\widetilde{B}_{t}}$ denote the extended belief state and its hidden states at turn $t$. The decoding of $\widetilde{B}_{t}$ (orange dashed box) is implemented in two different ways: (a) using a single decoder to generate the whole state, and (b) using two decoders to generate the domain-slot-value (DSV) triples and the topic separately.}
\label{hyknow}
\end{figure*}

\section{Task Definition}
In this section, we introduce our formulation of modeling TOD grounded on hybrid knowledge.
In particular, we assume that each dialog turn in TOD is grounded on either structured or unstructured knowledge.
We formulate the modeling of the two kinds of dialog turns separately.

In turns that are grounded on structured knowledge, the system needs to track user goals, i.e. the belief state, as domain-slot-value triples, and then query a database (DB) to guide response generation.
Specifically, we denote the user utterance and the system response at turn $t$ as $U_{t}$ and $R_{t}$ respectively.
Given the dialog context $C_{t}=[U_{t-k},R_{t-k},...,U_{t}]$ and previous belief state $B_{t-1}$, the system needs to generate current belief state $B_{t}$, which is formulated as $B_{t}=f^{(s)}_{b}(C_{t},B_{t-1})$.
Then the system performs DB query based on $B_{t}$ to get the matching result $\textbf{m}_{t}$. In this paper, we follow \citet{budzianowski2018multiwoz} to represent $\textbf{m}_{t}$ as a vector indicating the number of matched entities and whether the booking is available or not.
Afterwards, the system generates the response $R_{t}$, formulated as $R_{t}=f^{(s)}_{r}(C_{t},B_{t},\textbf{m}_{t})$.

In turns that are grounded on unstructured knowledge, the system manages a document base to guide response generation, which contains lists of documents characterized by different domains and entities, as showed in Fig.~\ref{sskm}.
Specifically, given the dialog context $C_{t}$, the system first retrieves a relevant document $D_{t}$ in the document base, formulated as $D_{t}=f^{(u)}_{d}(C_{t})$.
Then the system generates the response $R_{t}$ based on $C_{t}$ and retrieved $D_{t}$, which is formulated as $R_{t}=f^{(u)}_{r}(C_{t},D_{t})$. Noting that the original belief state is not updated in the unstructured knowledge-grounded turns, namely $B_{t}=B_{t-1}$. However, in this paper, we introduce extra belief state extension to facilitate the document retrieval.

\section{Proposed Framework}
Fig.~\ref{hyknow} shows an overview of our proposed system \ourmodel with end-to-end sequence-to-sequence (Seq2Seq) implementations.
It addresses our proposed task in three steps.
First, it uses the \textbf{extended belief tracking} to track user goals through dialog turns that involve hybrid knowledge.
Secondly, it performs \textbf{hybrid knowledge operations} based on the extended belief state, to search structured and unstructured knowledge that is relevant to the user goals.
Finally, it uses the extended belief state and relevant 
 knowledge to perform the \textbf{knowledge-grounded response generation}.

\subsection{Extended Belief Tracking}
\textbf{Belief State Extension.}
We define an extended belief state $\widetilde{B}_{t}$ which is applicable to track user goals in TODs that are grounded on both structured and unstructured knowledge.
Specifically, in turns that are grounded on structured knowledge, $\widetilde{B}_{t}$ is same as the original $B_{t}$, which describes user goals as domain-slot-value triples.
While in turns that are grounded on unstructured knowledge, $\widetilde{B}_{t}$ has an additional slot \textit{ruk} to indicate that current dialog turn \textbf{r}equires \textbf{u}nstructured \textbf{k}nowledge. The prefix and value of the slot \textit{ruk} represent the involved domain and entity, e.g. \textit{restaurant}-\textit{ruk}: \textit{Pizza Hut} colored in red in Fig.~\ref{sskm}.
We denote the combination of original and newly introduced domain-slot-value triples as $DSV_{t}$.
In addition, the \textit{topic} of $U_{t}$ is abstracted in $\widetilde{B}_{t}$ as a word sequence $T_{t}$ in each unstructured knowledge-grounded turn, e.g. \textit{favorite} colored in blue in Fig.~\ref{sskm}.

\textbf{Extended Belief State Decoding.}\label{sec:ebsd}
Following Seq2Seq framework, we first use the \textit{context encoder} to encode the dialog context $C_{t}$, whose last output is used as the initial hidden state of decoders.
Based on the hidden states of context encoder $\textbf{h}_{enc}^{C_{t}}$ and previous extended belief state $\textbf{h}_{enc}^{\widetilde{B}_{t-1}}$, we then decode the current extended belief state $\widetilde{B}_{t}$ under two schemes, which are described as below.

Since $DSV_{t}$ and $T_{t}$ are grounded on quite different vocabularies, we consider implementing the decoding of $\widetilde{B}_{t}$ in two ways: (a) using the belief state decoder to generate the whole $\widetilde{B}_{t}$, and (b) using the DSV decoder and the topic decoder to generate $DSV_{t}$ and $T_{t}$ separately.
Each implementation has its own advantages over the other.
Specifically, in the single-decoder implementation, the decoding of $DSV_{t}$ and $T_{t}$ can be jointly optimized via shared parameters:
\begin{align}
    \widetilde{B}_{t}=\mathrm{Seq2Seq}^{(b)}(C_{t}|\textbf{h}_{enc}^{\widetilde{B}_{t-1}}).
\end{align}
While in the multi-decoder implementation, the decoding of $DSV_{t}$ and $T_{t}$ are fitted to their own smaller decoding spaces (vocabularies), and thus the generation of $\widetilde{B}_{t}$ can be decomposed into two simpler decoding processes:
\begin{align}
\begin{split}
   DSV_{t}&=\mathrm{Seq2Seq}^{(dsv)}(C_{t}|\textbf{h}_{enc}^{\widetilde{B}_{t-1}}), \\
   T_{t}&=\mathrm{Seq2Seq}^{(t)}(C_{t}|\textbf{h}_{enc}^{\widetilde{B}_{t-1}}), \\
   \widetilde{B}_{t}&=[DSV_{t},T_{t}].
\end{split}
\end{align}

\subsection{Hybrid Knowledge Operations}
Based on the extended belief state $\widetilde{B}_{t}$, we conduct both DB query and document retrieval to get the query result $\textbf{m}_{t}$ and the relevant document $D_{t}$, which are used to guide the generation of response.
In the operation of DB query, we simply match the original triples in $\widetilde{B}_{t}$ with the DB entries.
While in the operation of document retrieval, we first preprocess the document base to extract the topic of each document as its retrieval index, e.g. \textit{vegetarian} and \textit{favorite} colored in yellow in Fig.~\ref{sskm}.
Then we use the extended part of $\widetilde{B}_{t}$ to match the domain, entity and extracted topic of each document, and select the best-matched one as $D_{t}$.\footnote{
See Appendix \ref{document} for more details of the document preprocessing and matching.}

\subsection{Knowledge-Grounded Response Generation}
We generate system response based on the dialog context $C_{t}$, the extended belief state $\widetilde{B}_{t}$, and the outputs of hybrid knowledge operations $\textbf{m}_{t}$ and $D_{t}$.
We first use the same context encoder in Sec.~\ref{sec:ebsd} to encode $C_{t}$.
Moreover, we use the \textit{belief state encoder} and the \textit{document encoder} to encode $\widetilde{B}_{t}$ and $D_{t}$ into hidden states $\textbf{h}_{enc}^{\widetilde{B}_{t}}$ and $\textbf{h}_{enc}^{D_{t}}$, respectively.
Based on the hidden states of all the encoders and the vector $\textbf{m}_{t}$, we use the \textit{response decoder} to generate the system response $R_{t}$, formulated as:
\begin{align}
\begin{split}
    \textbf{h}_{enc}^{\widetilde{B}_{t}}&=\mathrm{Encoder^{(b)}}(\widetilde{B}_{t}), \\
    \textbf{h}_{enc}^{D_{t}}&=\mathrm{Encoder^{(d)}}(D_{t}), \\
    R_{t}&=\mathrm{Seq2Seq}^{(r)}(C_{t}|\textbf{h}_{enc}^{\widetilde{B}_{t}},\textbf{h}_{enc}^{D_{t}},\textbf{m}_{t}),
\end{split}
\end{align}
where $\mathrm{Encoder^{(b)}}$ and $\mathrm{Encoder^{(d)}}$ denote the belief state encoder and the document encoder.

Following previous TOD systems with Seq2Seq architectures \cite{lei2018sequicity,liang2020moss,zhang2020probabilistic,zhang2020task}, we use one-layer, bi-directional GRU as encoders and standard GRU as decoders.
We also apply global attention \cite{bahdanau2015neural} and copy mechanism \cite{gu2016incorporating} in all the Seq2Seq processes, to improve the context-awareness of decoding $\widetilde{B}_{t}$ and $R_{t}$.

\subsection{Model Training}
\ourmodel is optimized through supervised training.
Specifically, each dialog turn in the training data is initially labeled with the original belief state and the relevant document.
We extend the belief state label based on the domain, entity and extracted topic of the relevant document.
Then the extended belief state label and the reference response are used to calculate the cross-entropy loss with the generated $\widetilde{B}_{t}$ and $R_{t}$, respectively.
We sum the two losses together and perform gradient descent in each turn to optimize the model parameters.
In our paper, the dialog context $C_t$ is set as the concatenation of previous system response $R_{t-1}$ and current user utterance $U_t$.\footnote{See Appendix \ref{implement} for more implementation details.}

\section{Experimental Settings}
\subsection{Dataset}
We evaluate our proposed system on the modified MultiWOZ 2.1 \cite{kim2020beyond} dataset, where crowd-sourcing workers are hired to insert additional turns into the original MultiWOZ dialogs. Each newly inserted turn is grounded on unstructured knowledge in one of the four domains: restaurant, hotel, taxi and train, with the label of its relevant document in the document base.
While the other three MultiWOZ domains (attraction, hospital and police) are not involved in these new turns.\footnote{See Appendix \ref{statistics} for details of data statistics.}

\begin{table*}[t]
\centering
\resizebox{0.9\textwidth}{!}{
\smallskip\begin{tabular}{lccccccc}
\hline
Model               & Pretrained LM &     Inform    &    Success    &      BLEU     &     METEOR    &     ROUGE-L   &    Combined   \\
\hline
UniConv             &      none     &      71.5     &      61.8     &      18.5     &      37.8     &      40.5     &      85.7     \\
LABES-S2S           &      none     &      76.5     &      65.3     &      17.8     &      36.8     &      39.9     &      88.7     \\
UniConv + BDA       &        -      &      72.0     &      62.6     &      16.9     &      35.7     &      38.9     &      84.2     \\
LABES-S2S + BDA     &        -      &      77.1     &      66.2     &      15.7     &      33.8     &      37.8     &      87.4     \\
\ourmodel (Single)      &      none     & \textbf{81.9} & \textbf{68.3} & \textbf{19.0} & \textbf{38.5} &      40.9     & \textbf{94.1} \\
~~~- w/o Joint Optim    &      none     &      78.5     &      65.7     &      18.3     &      36.9     &      39.6     &  90.4 (-3.7)  \\
\ourmodel (Multiple)    &      none     &      79.1     &      67.6     &      18.7     &      38.1     & \textbf{41.0} &      92.1     \\
~~~- w/o Joint Optim    &      none     &      77.7     &      65.4     &      18.0     &      36.6     &      39.5     &  89.6 (-2.5)  \\
\hline
SimpleTOD           &      GPT-2    &      81.7     &      67.9     &      14.5     &      34.2     &      37.0     &      89.3     \\
SimpleTOD + BDA     &        -      &      83.3     &      68.6     &      14.8     &      33.6     &      36.5     &      90.8     \\
\hline
\end{tabular}
}
\caption{End-to-end evaluation results on modified MultiWOZ 2.1. ``+" denotes the combination of Beyond Domain APIs (BDA) with E2E TOD models. Best results among light-weight systems (i.e. above internal dividing line) are marked in bold.
Evaluation metrics are described and marked in bold in Sec.~\ref{sec:e2e}.
}
\label{table1}
\end{table*}

\subsection{Baselines}
We compare \ourmodel with 1) existing end-to-end (E2E) TOD models and dialog state tracking (DST) models, to show the benefits of incorporating unstructured knowledge management into TOD modeling.
We also compare \ourmodel with 2) unstructured knowledge management models, to investigate our system's document retrieval performance.
For the comparison with pipeline systems that have hybrid knowledge management, we also consider the combinations of 1) and 2) as our baselines.

\textbf{E2E TOD Models and DST Models.}
We consider three baseline E2E TOD models with different types of structures: \textbf{UniConv} \cite{le2020uniconv} uses a structured fusion \cite{mehri2019structured} design, \textbf{LABES-S2S} \cite{zhang2020probabilistic} uses a multi-stage Seq2Seq \cite{lei2018sequicity} architecture, and \textbf{SimpleTOD} \cite{hosseini2020simple} is based on a single auto-regressive language model initialized from GPT-2 \cite{radford2019language}. All three E2E models only manage structured knowledge (database) in their TOD modeling.
In addition to E2E TOD models, we also compare \ourmodel with existing DST models in the belief tracking evaluation.
Specifically, we use \textbf{TRADE} \cite{wu2019transferable} and \textbf{TripPy} \cite{heck2020trippy} as two DST baselines, which are representative BERT-free and BERT-based DST models, respectively.

\textbf{Unstructured Knowledge Management Models.}
We first compare our system with Beyond Domain APIs (\textbf{BDA}) \cite{kim2020beyond}.
This baseline model uses two classification modules based on BERT \cite{devlin2019bert} to detect unstructured knowledge-grounded dialog turns and retrieve relevant documents, respectively.
Moreover, we use standard information retrieval (IR) systems \textbf{TF-IDF} \cite{manning2008introduction} and \textbf{BM25} \cite{robertson2009probabilistic} as the other two baseline models.

\textbf{Combinations.}
We combine the unstructured knowledge management model BDA with every DST or E2E TOD model. Specifically, BDA detects dialog turns that are grounded on unstructured knowledge, and uses a fine-tuned GPT-2 to generate responses in these turns, based on the dialog context and retrieved documents. While the DST or E2E TOD model handles the rest dialog turns which are grounded on structured knowledge.

Noting that TripPy and SimpleTOD use large-scale pretrained language models (LM) to improve their dialog modeling performance, which requires large model sizes and computing resources.
For fair comparisons, we distinguish them from other light-weight models in our experiments.

\section{Results and Analysis}
We test our system's performance under both the single-decoder and multi-decoder belief state decoding implementations, denoted as \ourmodel (Single) and \ourmodel (Multiple), respectively.
Both implementations of \ourmodel come to the same conclusions when compared with the baseline models, which are described in detail below.

\subsection{End-to-End Evaluation}\label{sec:e2e}
Table \ref{table1} shows our experimental results of the end-to-end (E2E) evaluation, where we evaluate the task completion rate and language quality of system responses.
In terms of the task completion rate, we measure whether the system provides correct entities (\textbf{Inform} rate) and answers all the requested information (\textbf{Success} rate) in a dialog, following \citet{budzianowski2018multiwoz}.
For the evaluation of language quality, we adopt commonly used metrics \textbf{BLEU} \cite{papineni2002bleu}, \textbf{METEOR} \cite{banerjee2005meteor} and \textbf{ROUGE-L} \cite{lin2004rouge}.
Moreover, we use \textbf{Combined} score computed by $(Inform+Success)\times0.5+BLEU$ for overall evaluation, as suggested by \citet{eric2020multiwoz}.

We find that \ourmodel has better task completion rate than the light-weight E2E TOD models, which is comparable with SimpleTOD who uses large-scale pretrained GPT-2.
It also generates responses with better language quality compared to all the E2E models.
This is because our extended belief state can distinguish whether a dialog turn is grounded on structured or unstructured knowledge, which avoids the confusion between handling the two kinds of turns.
In addition, we manage the document base to provide relevant references for generating the response, which guide our system to give more appropriate responses in turns that are grounded on unstructured knowledge.

We also observe that \ourmodel outperforms the combinations of BDA and light-weight E2E TOD model.
This indicates that our end-to-end model framework has advantages over the pipeline structures of combination models.
In particular, dialog modeling grounded on the structured and unstructured knowledge are integrated in a uniform Seq2Seq architecture in our system, where they are jointly optimized to an overall better performance.
Although \ourmodel does not significantly outperform the combination of BDA and SimpleTOD, our system has lower deployment cost since it is trained end-to-end.

\begin{table}[t]
\centering
\resizebox{1.0\columnwidth}{!}{
\smallskip\begin{tabular}{l@{~~}c@{~~}c@{~~}c@{~~}c}
\hline
Model               &     Inform    &    Success    &      BLEU     &    Combined    \\
\hline
UniConv             &      84.2     &      71.8     &      19.0     &      97.3      \\
LABES-S2S           &      83.6     &      74.2     &      18.3     &      97.2      \\
UniConv + BDA       &      85.8     &      73.9     &      19.3     &      99.4      \\
LABES-S2S + BDA     &      85.0     &      75.3     &      18.9     &      99.1      \\
\ourmodel               & \textbf{87.2} & \textbf{76.5} & \textbf{19.5} & \textbf{101.4} \\
\hline
SimpleTOD           &      87.5     &      76.4     &      16.3     &      98.3      \\
SimpleTOD + BDA     &      89.0     &      77.2     &      17.0     &      100.1     \\
\hline
\end{tabular}
}
\caption{Context-to-response generation results on modified MultiWOZ 2.1. All symbols and markings have the same meaning as in Table \ref{table1}.}
\label{table2}
\end{table}

\subsection{Context-to-Response Generation}
We also conduct evaluations on the context-to-response (C2R) generation, where systems directly use the oracle belief state and knowledge to generate the response.
The experimental results are shown in Table \ref{table2}, where we observe the same conclusions as in the E2E evaluation (Table \ref{table1}). This again shows our system's superiority in TOD modeling grounded on hybrid knowledge.

Additionally, we observe that \ourmodel's performance gap between E2E and C2R evaluations is smaller than the baseline models, reflected in the smaller variations of the combined score.
This shows that the belief state and knowledge provided by our system are probably closer to the oracle and may give stronger guidance to generate a response.

\begin{table}[t]
\centering
\resizebox{0.9\columnwidth}{!}{
\smallskip\begin{tabular}{lcc}
\hline
Model               & Pretrained LM &   Joint Goal  \\
\hline
TRADE               &      none     &      42.9     \\
UniConv             &      none     &      45.5     \\
LABES-S2S           &      none     &      46.0     \\
TRADE + BDA         &        -      &      43.8     \\
UniConv + BDA       &        -      &      46.5     \\
LABES-S2S + BDA     &        -      &      46.8     \\
\ourmodel (Single)      &      none     & \textbf{48.0} \\
~~~- w/o Joint Optim   &      none     &  46.2 (-1.8)  \\
\ourmodel (Multiple)    &      none     &      47.6     \\
~~~- w/o Joint Optim   &      none     &  45.6 (-2.0)  \\
\hline
TripPy              &      BERT     &      50.4     \\
SimpleTOD           &      GPT-2    &      48.4     \\
TripPy + BDA        &        -      &      51.2     \\
SimpleTOD + BDA     &        -      &      49.8     \\
\hline
\end{tabular}
}
\caption{Original turns' belief tracking results on modified MultiWOZ 2.1. ``+" denotes the combination of BDA with DST/E2E models. The best result among light-weight systems (i.e. above internal dividing line) is marked in bold.
The evaluation metric is described and marked in bold in Sec.~\ref{sec:km}.
}
\label{table3}
\end{table}

\subsection{Knowledge Management}\label{sec:km}
To further investigate our system's end-to-end performance, we conduct evaluations on the intermediate knowledge management.
In particular, we evaluate the structured and unstructured knowledge management separately in the original and newly inserted dialog turns.
In the original turns grounded on structured knowledge, we evaluate the belief tracking performance which directly determines the database query accuracy.
Specifically, we use the \textbf{Joint Goal} accuracy \cite{henderson2014second} to measure whether belief states are predicted correctly in a dialog turn.
While in newly inserted turns grounded on unstructured knowledge, we adopt standard information retrieval metrics \textbf{R@1} and \textbf{MRR@5} to evaluate the document retrieval performance.
Table \ref{table3} and \ref{table4} shows our evaluation results of belief tracking and document retrieval, respectively.

\begin{table}[t]
\centering
\resizebox{1.0\columnwidth}{!}{
\smallskip\begin{tabular}{l@{~~}c@{~~}c@{~~}c@{~~}c}
\hline
Model                   &      Type       &     MRR@5     &     R@1       \\
\hline
TF-IDF                  &   standard IR   &     68.7      &     54.1      \\
BM25                    &   standard IR   &     69.2      &     52.5      \\
BDA                     & classification  &     80.6      &     69.8      \\
\hline
\ourmodel (Single)           & topic match  & \textbf{81.7} & \textbf{80.2} \\
~~~- w/o Joint Optim         & topic match  &  80.1 (-1.6)  &  77.8 (-2.4)  \\
\ourmodel (Multiple)         & topic match  &      81.1     &      79.5     \\
~~~- w/o Joint Optim         & topic match  &  79.7 (-1.4)  &  77.4 (-2.1)  \\
\hline
\end{tabular}
}
\caption{Newly inserted turns' document retrieval results on modified MultiWOZ 2.1.
Best results are marked in bold.
Evaluation metrics are described and marked in bold in Sec.~\ref{sec:km}.
}
\label{table4}
\end{table}

In terms of belief tracking, \ourmodel outperforms the light-weight DST/E2E models. 
This is because our extended belief tracking can detect the newly inserted turns apart from the original turns (via the slot \textit{ruk}), which improves our system's awareness on deciding when to update the original triples in the belief state. 
\ourmodel also has better belief tracking performance than the combinations of BDA and light-weight DST/E2E model.
This is because error propagation on updating belief states is eliminated in our system compared to the pipeline framework: The pipeline system either updates the belief state or retrieves the document in one turn, but \ourmodel can perform both operations in the nature of its E2E design.
Although the belief tracking performance of \ourmodel is not as good as that of TripPy and SimpleTOD, our system does not use large-scale pretrained BERT or GPT-2 and is thus computational cheaper.\footnote{See Appendix \ref{size} for details on model size comparison.}

In the document retrieval evaluation, we find that \ourmodel outperforms the unstructured knowledge management models, especially on the R@1 metric.
This shows that our system's document retrieval scheme with topic matching has a higher accuracy, compared to the classifier-based BDA and the standard information retrieval (IR) systems.
Specifically, \ourmodel retrieves documents based on the highly simplified semantic information, i.e. the topic, which reduces the complexity of the retrieval process.
This makes the retrieval scheme of \ourmodel more concise and effective than the baseline models, who directly calculate the relevance of dialog context to every document content.

\subsection{Single vs. Multiple Decoders}
We then compare our two implementations of extended belief state decoding. We calculate the vocabularies of DSV triples, the topic and their combination (which are 709, 166 and 862), and observe that the last one approximately equals to the sum of the former two.
This confirms our assumption in Sec.~\ref{sec:ebsd} that DSV triples and topic have quite different vocabularies, which motivates the multi-decoder implementation in belief state decoding.

However, we find that \ourmodel (Single) outperforms \ourmodel (Multiple) in both E2E and knowledge management evaluations, as shown in Table \ref{table1}, \ref{table3} and \ref{table4}.
This shows that the decoding of DSV triples and topic can benefit from the joint optimization via shared parameters, although they are grounded on quite different vocabularies.
The superiority of joint optimization further implies that the structured and unstructured knowledge management in TOD modeling have a positive correlation, since they commonly involve task-specific domain knowledge and entities. Therefore, the two kinds of knowledge management can learn from each other through joint training, and achieve overall better performance compared to separating them apart.

\begin{table*}[t]
\centering
\resizebox{0.9\textwidth}{!}{
\smallskip\begin{tabular}{clccccccc}
\hline
        Test Set          & Model                  &  Joint Goal  & Inform & Success &  BLEU  & METEOR & ROUGE-L &  Combined  \\
\hline
\multirow{3}{*}{Original} & LABES-S2S + BDA        &     49.0     &  82.1  &  69.8   &  17.8  &  37.1  &  40.2   &    93.8    \\
                          & SimpleTOD + BDA        &     51.8     &  85.6  &  70.9   &  16.3  &  34.5  &  38.6   &    94.6    \\
                          & \ourmodel (Single)     &     49.2     &  82.3  &  69.4   &  18.0  &  37.3  &  40.2   &    93.9    \\
\hline
\multirow{3}{*}{Modified} & LABES-S2S + BDA        &  46.8 (-2.2) &  77.1  &  66.2   &  17.7  &  36.8  &  39.6   &  89.4 (-4.4) \\
                          & SimpleTOD + BDA        &  49.8 (-2.0) &  83.3  &  68.6   &  15.8  &  33.6  &  37.8   &  91.8 (-2.8) \\
                          & \ourmodel (Single)     &  48.0 (-1.2) &  81.9  &  68.3   &  17.8  &  37.2  &  39.5   &  92.9 (-1.0) \\
\hline
\end{tabular}
}
\caption{End-to-end evaluation results on the original and modified MultiWOZ 2.1 test set. The evaluation is conducted only in the original dialog turns.}
\label{table5}
\end{table*}

\subsection{Ablation Study}
We ablate the joint optimization of structured and unstructured knowledge-grounded TOD modeling to investigate its role in our framework, denoted as w/o Joint Optim in Table \ref{table1}, \ref{table3} and \ref{table4}.
Specifically, we train two \ourmodel models separately on the original and newly inserted dialog turns, and use them to handle TOD grounded on structured and unstructured knowledge, respectively.
To determine which model should be used, the oracle label of slot \textit{ruk} is used to judge which knowledge type the current dialog turn is grounded on.

We observe that removing joint optimization brings \ourmodel evident performance declines in the end-to-end evaluation (Table \ref{table1}). This suggests that joint optimization plays a significant role in improving \ourmodel's end-to-end performance, where TOD modeling grounded on the two kinds of knowledge can benefit each other by learning shared parameters.
The ablation of joint optimization also causes performance declines in \ourmodel's knowledge management (Table \ref{table3} and \ref{table4}). This again indicates that the two kinds of knowledge management are positively correlative and can get benefit from joint training.

\subsection{Between Structured and Unstructured Knowledge}
In this section, we investigate how the newly inserted dialog turns (grounded on unstructured knowledge) affect systems' E2E performance in the original dialog turns (grounded on structured knowledge).
Specifically, we evaluate systems' E2E performance on both the original and modified MultiWOZ 2.1 test sets.
This evaluation is conducted only in the original dialog turns, which is different from the E2E evaluation conducted in all turns (Table \ref{table1}).
Table \ref{table5} shows the results of this experiment, where we compare \ourmodel (Single) with strong combination models.

We find that all the models' performance is degraded when transferred from the original to the modified test set. This indicates that the inserted turns grounded on new knowledge may interrupt the original dialogs, which complicates the dialog process and causes difficulties in the original turns' dialog modeling.

However, we observe that \ourmodel (Single) suffers from less reduction compared to the baseline combination models.
This shows that our system has a stronger resistance to the interruptions of newly inserted turns, which benefits from our end-to-end modeling.
Specifically, \ourmodel jointly optimizes dialog modeling of the original and newly inserted turns in a uniform end-to-end framework. This unified modeling approach improves our system's flexibility in switching between the two kinds of turns, and thus makes it more competent in handling the complicated dialog process.

\begin{table}[t]
\centering
\resizebox{1.0\columnwidth}{!}{
\smallskip\begin{tabular}{@{~}l@{~~}c@{~~}c@{~~}cc@{~~}c@{~~}c@{~}}
\hline
\multirow{2}{*}{Model} & \multicolumn{3}{c}{Original}                & \multicolumn{3}{c}{Newly Inserted}         \\
                         \cmidrule(lr){2-4}                              \cmidrule(lr){5-7}              
                       &      Cohe.     &      Info.     &     Corr.      &      Cohe.     &      Info.     &      Corr.     \\
\hline
SimpleTOD              &      2.58     &      2.56     &     2.44      &      2.50     &      2.04     &      2.14     \\
SimpleTOD + BDA        &      2.56     &      2.60     & \textbf{2.46} &      2.52     &      2.30     &      2.22     \\
\hline
\ourmodel (Single)     & \textbf{2.60} & \textbf{2.62} &     2.42      & \textbf{2.56} & \textbf{2.36} & \textbf{2.50} \\
\hline
\end{tabular}
}
\caption{Human evaluation results on modified MultiWOZ 2.1, results in original and newly inserted turns are shown separately.}
\label{table6}
\end{table}

\begin{table*}[t]
\centering
\small
\resizebox{1.0\textwidth}{!}{
\begin{tabular}{c|l}
\hline
\multicolumn{2}{l}{$U_{1}$: Hello, I would like to find a \textbf{hotel} that has WiFi in the north part of the town.} \\
\multicolumn{2}{l}{$B_{1}$: hotel-area: north, hotel-internet: yes} \\
\multicolumn{2}{l}{$R_{1}$: The \textbf{Arbury Lodge Guesthouse} is one of 12 options for you. Shall I make a reservation for you?} \\
\multicolumn{2}{l}{$U_{2}$: Do they provide \textbf{Italian breakfast}?} \\
\hline
\multirow{3}*{Human} & $B_{2}$: hotel-area: north, hotel-internet: yes \\
                     & $D_{2}$: Full English \textbf{breakfast} is served every morning at Arbury Lodge Guesthouse. \\
                     & $R_{2}$: \textbf{No}, but they have a full English breakfast. Did you still want to book a room here? \\
\hline
\multirow{2}*{SimpleTOD} & $B_{2}$: hotel-area: north, hotel-internet: yes, \textbf{restaurant-food: Italian} \\
                     & $R_{2}$: Da Vinci Pizzeria serves Italian food and is located in the north. \\
\hline
\multirow{3}*{SimpleTOD + BDA} & $B_{2}$: hotel-area: north, hotel-internet: yes \\
                    & $D_{2}$: German, Italian, Spanish and English are \textbf{spoken} at Carolina Bed and Breakfast. \\
                    & $R_{2}$: \textbf{Yes}, the hotel has a Italian breakfast. Would you like to make a reservation?\\
\hline
\multirow{3}*{\ourmodel (Single)} & \\[-9pt]
                    & $\widetilde{B}_{2}$: hotel-area: north, hotel-internet: yes, \textbf{hotel-ruk: Arbury Lodge Guesthouse}, \textbf{breakfast}\\
                    & $D_{2}$: Full English \textbf{breakfast} is served every morning at Arbury Lodge Guesthouse. \\
                    & $R_{2}$: \textbf{No}, they provides full English breakfast. Would you like me to book a room for you?\\
\hline
\end{tabular}
}
\caption{Comparison of the belief tracking, document retrieval and response generation results when dialog steps into a newly inserted turn of modified MultiWOZ 2.1. ``Human'' denotes ground truth results.}
\label{table7}
\end{table*}

\subsection{Human Evaluation}
There is still a gap between the evaluation results of automatic metrics and the real E2E performance of TOD systems.
Therefore, we conduct human evaluation to more adequately test our system's E2E performance. In particular, we compare \ourmodel (Single) with a strong E2E baseline SimpleTOD and its combination with BDA.

We conduct human evaluation separately on the two types (original and newly inserted) of dialog turns.
Specifically, we sample fifty dialog turns of each type and ask the judges to evaluate each turn's system response on three aspects.
\textbf{Coherence} (Cohe.) measures how well the response is coherent with the dialog context.
\textbf{Informativeness} (Info.) measures how well the response can provide sufficient information that meets the user requests.
\textbf{Correctness} (Corr.) measures how well the information in response is consistent with the ground truth knowledge, i.e. relevant DB entries or documents.
All the three aspects are scored on a Liker scale of 1-3, which denotes \textit{bad}, \textit{so-so} and \textit{good}.

Table \ref{table6} shows our human evaluation results.
In the original dialog turns, \ourmodel (Single) scores close to SimpleTOD and its combination with BDA on all the three aspects. This indicates that our proposed light-weight system is comparable with the large GPT-2 based models in managing structured knowledge to generate the response.
In addition, our model outperforms the two baseline models in the newly inserted dialog turns.
Specifically, \ourmodel (Single) generates responses with significantly better informativeness and correctness than SimpleTOD. This again shows that the management of unstructured knowledge is beneficial for generating appropriate responses.
Compared to the combination of SimpleTOD and BDA, the responses generated by \ourmodel (Single) also achieve much better correctness, which benefits from our model’s higher document retrieval accuracy (as shown in Table \ref{table4}).

\subsection{Case Study}
An example dialog segment ($U_1$, $B_1$, $R_1$, $U_2$) and corresponding output results of each model ($B_2$, $D_2$, $R_2$) are presented in Table \ref{table7}.
Without access to unstructured document base, SimpleTOD misunderstands the user query, and instead recognizes the term ``Italian'' in user utterance as a constraint to update the belief state. As a result, the system makes an inappropriate recommendation.
By combining with BDA, SimpleTOD predicts correct belief state, but fails in finding the relevant document, thus providing a wrong answer. This is because the wrong document’s content has many common words with the dialog context, e.g. “Italian” and “Breakfast”, which mislead the retrieval process of BDA.
In contrast, \ourmodel gives a proper response with accurate information as it identifies the entity (``Arbury Lodge Guesthouse'') and captures the topic (``breakfast'') to avoid the misleading of common words in document retrieval.

\section{Conclusion}
In this paper, we define a task of modeling TOD with access to both structured and unstructured knowledge.
To address this task, we propose a TOD system \ourmodel which uses an E2E framework to jointly optimize TOD modeling grounded on the two kinds of knowledge.
In the experiments, \ourmodel shows strong performance in modeling TOD with hybrid knowledge management, compared to existing TOD systems and their pipeline extensions.
For future work, we plan to incorporate large-scale pretrained language models into our proposed system to further enhance its performance. Furthermore, we consider evaluating our system on different scenarios where dialogs are grounded on hybrid knowledge.

\section*{Acknowledgments}
This work was partly supported by the NSFC projects (Key project with No.~61936010 and regular project with No.~61876096).
This work was also supported by the Guoqiang Institute of Tsinghua University, with Grant No.~2019GQG1 and 2020GQG0005.
We would like to thank colleagues from HUAWEI for their constant support and valuable discussion.

\bibliographystyle{acl_natbib}
\bibliography{acl2021}

\newpage
\appendix
\section{Document Preprocessing and Matching}
\label{document}
We preprocess the document base of the modified MultiWOZ 2.1 \cite{kim2020beyond} dataset to extract the topic of each document, which are used as its retrieval index in the unstructured knowledge management.
Based on the TF-IDF \cite{manning2008introduction} algorithm, we perform the topic word extraction domain-by-domain in a two-step procedure.
First, we choose the top-three keywords with the highest TF-IDF scores in each document as its topic candidates.
Then we filter the candidates to further select our desired topic words.

Noticing that different entities in the same domain usually have documents covering similar topics, we assume that a desired topic word should typically appear in multiple entities' documents, and therefore have a high frequency of occurrence among the topic candidates.
So we calculate a cumulative average TF-IDF (CA-TF-IDF) score for each topic word in the candidates, which synthetically measures the word's document-level TF-IDF and entity-level occurrence frequency.
Specifically, CA-TF-IDF sums the TF-IDF score of a topic word's each occurrence in the candidates, and divides it by the entity number in the domain.
We filter out the topic candidates with low CA-TF-IDF scores and retain the rest to form the final retrieval indexes.
The filtering thresholds are 2.3, 2.7, 6.9 and 7.3 for the domain of restaurant, hotel, taxi and train, respectively. While other domains are not involved in the document base.
After the preprocessing, each document has one to three topic words extracted.

In the document retrieval process, we use the prefix and value of slot \textit{ruk} in our proposed extended belief state to locate the document list of involved domain and entity.
Then we use the topic of user utterance in our extended belief state to match the extracted topic of each document in the involved list, and select the best-matched one as the relevant reference.
The topic matching is conducted by using the fuzzy string matching toolkit\footnote{\url{https://github.com/seatgeek/fuzzywuzzy}}.
Noting that the relevant document is set to \textit{none} if the slot \textit{ruk} or the topic of user utterance is not available.

\section{Implementation Details}
\label{implement}
We use GloVe \cite{pennington2014glove} to initialize the embedding matrix, and set batch size, embedding size, hidden size and vocabulary size as 40, 50, 200 and 3000, respectively.
We also set dropout rate as 0.35 and use greedy decoding to generate the belief state and response.
Moreover, we use Adam optimizer \cite{kingma2014adam} with a learning rate of $7e^{-4}$, which is selected via grid search from $\{4e^{-4},5e^{-4},6e^{-4},7e^{-4},8e^{-4},9e^{-4},1e^{-3}\}$.
We halve the learning rate when no improvement of overall performance (combined score) is observed on the development set in two consecutive epochs, and we stop the training when no improvement is observed in four consecutive epochs.
The average training time is about 80 minutes per epoch, and the total number of training epoch is around 15.
Model training is performed on NVIDIA TITAN-Xp GPU.

\section{Statistics of Modified MultiWOZ 2.1}
\label{statistics}
There are totally 8449/1001/1004 dialogs\footnote{These are slightly more compared to the original MultiWOZ 2.1, because some of the original dialogs are modified twice with different turns inserted.} in the training, development and testing set of modified MultiWOZ 2.1, where 6501/836/847 dialogs have new turns inserted, respectively.
After the modification, each dialog has 8.93 turns on average, which is longer than the original 6.85.
The ontology of modified MultiWOZ 2.1 is the same as the original, with 32 slot types (excluding \textit{ruk}) and 2426 corresponding slot values.

\begin{table}[t]
\centering
\resizebox{0.9\columnwidth}{!}{
\smallskip\begin{tabular}{lcc}
\hline
Model                & Pretrained LM &   Model Size \\
\hline
TRADE                &      none     &      10M     \\
UniConv              &      none     &      16M     \\
LABES-S2S            &      none     &      3.8M    \\
\ourmodel (Single)   &      none     &      4.1M    \\
\ourmodel (Multiple) &      none     &      5.3M    \\
\hline
TripPy               &      BERT     &      110M    \\
SimpleTOD            &      GPT-2    &      81M     \\
\hline
\end{tabular}
}
\caption{Comparison of model size.}
\label{appendixtable1}
\end{table}

\section{Model Size Comparison}
\label{size}
Table \ref{appendixtable1} shows the model size of our proposed \ourmodel and some baseline models.
We find that \ourmodel has a comparable model size with the light-weight baseline models, which do not leverage pretrained language models (LM).
But its model size is much smaller than that of TripPy and SimpleTOD, which use pretrained BERT and GPT-2, respectively.
Therefore, \ourmodel requires much less computational resources, compared to TripPy and SimpleTOD that use large-scale pretrained LM.

\end{document}